%% file: main.tex
\documentclass{ifacconf}
\usepackage{times,url}
\usepackage{graphicx,natbib,amsmath,amssymb,subfiles}
\graphicspath{{Figures2/}{../Figures2/}{figures/}{../figures/}}

\newtheorem{assumption}{Assumption}
\newcommand{\state}{\ensuremath{\mathit{X}}}

\newcommand{\dynamicA}{\ensuremath{\mathit{A}_*}}
\newcommand{\controlB}{\ensuremath{\mathit{B}_*}}
\newcommand{\extendedmatrix}{\ensuremath{\Theta}_*}
\newcommand{\actionU}{\ensuremath{\mathit{U}}}

\newcommand{\statecost}{\ensuremath{\mathit{Q}_x}}
\newcommand{\actioncost}{\ensuremath{\mathit{Q}_u}}
\newcommand{\dimstate}{\ensuremath{p}}
\newcommand{\dimaction}{\ensuremath{q}}
\newcommand{\dimnoise}{\ensuremath{r}}

\newcommand{\WeinerPP}{\ensuremath{\mathit{\mathbb{W}}}}
\newcommand{\numberofpieces}{\ensuremath{\mathit{M}}}
\newcommand{\lengthofdiscrete}{\ensuremath{\epsilon}}
\newcommand{\normalinput}{\ensuremath{\mathit{\mathbb{U}}}}
\newcommand{\sampleparameter}{\ensuremath{\widehat{\Theta}}}

\begin{document}
\begin{frontmatter}
\title{On Stochastic Continuous-Time Linear System Identification for Adaptive Stabilization}
\author[smu]{Reza Sadeghi Hafshejani}
\author[smu]{Mohamad Kazem Shirani Faradonbeh}
\address[smu]{Department of Mathematics, Southern Methodist University, Dallas, TX 75205, USA (e-mail: rezas@smu.edu; mohamadksf@smu.edu).}
\begin{abstract}

System identification studies the canonical problem of reliably learning system-dynamics from state and input trajectories. One of its important applications consists of estimating a feedback gain matrix that can stabilize an unknown system. This paper focuses on the analysis of such adaptive stabilization schemes for multivariate linear time-invariant state-space models. The system instability practically renders trajectories short, i.e., the observed data is notably limited. So, it is crucial to tightly consider effects of \emph{all} involved quantities including length and number of trajectories, dimension, controllability Gramian, and signal-to-noise ratio. We put forward a simple, practical, and reliable relationship between the chance that adaptive stabilization (based on system identification) fails to stabilize the system, in terms of all the above-mentioned quantities. Our findings show that despite exponential effects of all other quantities on failure probability, such effects do not hold for signal-to-noise ratio. More specifically, this study shows that the strength of the excitation signal helps stabilization only \emph{logarithmically}; a very important phenomenon that we explain under the term \emph{signal saturation}.

\end{abstract}
\end{frontmatter}
\subfile{sections/01_introduction}

\subfile{sections/02_problem}

\subfile{sections/03_identification}

\subfile{sections/04_numerics}

\subfile{sections/05_deficiencies}
\bibliography{references}
\end{document}

%% file: sections/01_introduction.tex
\section{Introduction}\label{sec:introduction}

Linear control systems that evolve in continuous time are ubiquitous models, with applications ranging from robotics and aerospace to algorithmic trading \citep{treven2021stabilizing, faradonbeh2023online}. A central control problem is stabilization: choosing the control input so that the closed-loop dynamics are stable and the system can operate for extended periods \citep{treven2021stabilizing,lale2022fast,hu2022stabilizing}. When the system dynamics are known, methods for designing stabilizing feedback controllers are well established \citep{astrom2008feedback}. However, under unknown dynamics, a controller must instead be learned from observations of the system.

Adaptive control in unknown environments attracts significant attention, with feedback controllers learned through interaction with the environment to achieve suitable stability or optimality properties. In discrete time, non-asymptotic system identification is studied extensively. Several algorithms assume access to an initial stabilizing gain and aim to improve performance with respect to a cost function \citep{dean2018regret,simchowitz2020naive}. However, stabilization itself is nontrivial, and such prior knowledge cannot always be assumed \citep{hu2022stabilizing}. Accordingly, several algorithms are developed that \emph{learn to stabilize} without a given stabilizer.

These approaches include random linear feedbacks with high-probability stabilization guarantees \citep{faradonbeh2019stabilization}, robust synthesis from estimated uncertainty sets \citep{treven2021stabilizing}, and exploration strategies combining fast stabilization with low regret \citep{lale2022fast}. The sample complexity, however, depends on system structure, as weak controllability can substantially increase the data requirements for successful stabilization \citep{tsiamis2022hard}. Nevertheless, identifying the full dynamics need not be necessary. For noise-free systems under suitable spectral and controllability assumptions, learning the dynamics associated with the unstable subspace can suffice for stabilization, with sample complexity governed primarily by the number of unstable modes \citep{hu2022stabilizing}. This suggests favorable data requirements for systems with many stable modes and relatively few unstable ones.

In continuous time, non-asymptotic results are comparatively fewer. Existing approaches either assume an initial stabilizing controller \citep{faradonbeh2022thompson,faradonbeh2023online}, or use multiple finite-horizon realizations \citep{basei2022logarithmic,zhou2025finite}. A related Bayesian approach learns a stabilizing controller from a single trajectory, supported by numerical experiments and sufficient estimation-accuracy conditions for stability \citep{faradonbeh2022bayesian}. However, the joint effects of trajectory count, observation horizon, dimension, noise intensity, and input design on stabilization success remain insufficiently understood.

We study these effects numerically, examining how well continuous-time stochastic linear systems with possibly unstable dynamics can be stabilized from data, without an initial stabilizing feedback gain. We present a method that \emph{learns to stabilize} the system without prior knowledge of the dynamics, from either a single trajectory or multiple trajectories. We then examine how its failure probability depends on the aggregate observation time, dimension, disturbance intensity, input strength, and controllability. The experiments indicate that, under suitable excitation and controllability, the failure probability decays exponentially with the aggregate observation time. Thus, fewer longer trajectories can perform comparably to more shorter ones. This holds despite the challenges induced by instability, such as the exponential growth of the state and the strong statistical dependence along each trajectory.

The experiments also reveal a phenomenon that we call \emph{signal saturation}. Existing guarantees improve with the signal-to-noise ratio (SNR) between excitation and disturbance \citep{ziemann2023tutorial}. In discrete time, the least-squares error bound of \citet{dean2020sample} scales inversely with the SNR, implying that the probability that the resulting robust controller fails to stabilize the system decays exponentially in it. A similar rate holds for estimating the drift matrix in continuous time \citep{hafshejani2025effect}. In contrast, we observe that stronger excitation reduces the logarithm of the failure probability only logarithmically in the excitation variance rather than linearly, so its benefits saturate.

We also distinguish identification accuracy from stabilization success. Motivated by the unstable-subspace perspective of \citet{hu2022stabilizing}, we modify the input matrix to reduce excitation of a stable mode. Identification accuracy then degrades substantially, while stabilization performance remains essentially unchanged. This suggests that not every component of the dynamics is equally important for stabilization. Finally, we examine resilience to model-order misspecification. Our aim is to provide practitioners with a method for stabilization, with experimental evidence on the data and excitation requirements. A theoretical analysis is beyond the scope of this study.

The remainder of the paper is organized as follows. Section~\ref{sec:formulation} formulates the problem. Section~\ref{sec:identification} presents the method and the scaling relation, which Section~\ref{sec:numerics} examines numerically. Section~\ref{sec:conclusion} concludes the paper.

\textit{Notation.} For a matrix $A$, $A^\top$ denotes its transpose, $\lVert A\rVert$ its operator norm, $\lVert A\rVert_F$ its Frobenius norm, and $\lambda_{\max}(A)$ an eigenvalue of $A$ with largest real part. For symmetric matrices, $\lambda_{\min}(A)$ denotes the smallest eigenvalue, and $A\succeq B$ means that $A-B$ is positive semidefinite. We write $\mathbb{I}_m$ for the $m\times m$ identity matrix, and $\Re\{z\}$ for the real part of $z\in\mathbb{C}$. Finally, $\mathbb{N}(\mu,\Sigma)$ denotes a normal distribution with mean $\mu$ and covariance $\Sigma$.

%% file: sections/02_problem.tex
\section{Problem Formulation}\label{sec:formulation}
We consider a dynamical system whose state $\state_t\in\mathbb{R}^{\dimstate}$ evolves according to the stochastic differential equation (SDE)
\begin{equation}\label{eq:sde}
	d\state_t=\bigl(\dynamicA\state_t+\controlB\actionU_t\bigr)\,dt+C_t\,d\WeinerPP_t,
\end{equation}
where $\actionU_t\in\mathbb{R}^{\dimaction}$ is an input chosen by the controller at time $t$ and $\WeinerPP_t\in\mathbb{R}^{\dimnoise}$ is a standard Wiener process that disturbs the system. Here, $\dynamicA\in\mathbb{R}^{\dimstate\times\dimstate}$ is the system matrix governing the dynamics and $\controlB\in\mathbb{R}^{\dimstate\times\dimaction}$ is the input matrix that determines the effect of the control input on the state. Moreover, $C_t\in\mathbb{R}^{\dimstate\times\dimnoise}$ is a time-varying noise coefficient matrix that satisfies
\begin{equation}\label{eq:upper-bound-C}
	\lVert C_t\rVert\le \sigma_{\mathbb{W}},\qquad\text{for all}\qquad t\ge0,
\end{equation}
for some constant $\sigma_{\mathbb{W}}>0$.

All matrices in the above setting are assumed to be \emph{unknown}. Moreover, we do not impose any assumption on the eigenstructure of $\dynamicA$, so the open-loop dynamics may be unstable. The pair $(\dynamicA,\controlB)$ is said to be stabilizable if there exists a feedback gain $K$ such that all eigenvalues of $\dynamicA+\controlB K$ lie strictly in the open left half-plane of the complex plane. 

For a stabilizable pair $(A,B)$, let $\Theta^\top=[A\;B]$ denote their horizontal concatenation. For positive definite matrices $\statecost\in\mathbb{R}^{\dimstate\times\dimstate}$ and $\actioncost\in\mathbb{R}^{\dimaction\times\dimaction}$, let $P(\Theta)$ be the symmetric positive definite solution of the Riccati equation
\begin{equation}\label{eq:care}
	A^\top P(\Theta)+P(\Theta)A-P(\Theta)B\actioncost^{-1}B^\top P(\Theta)+\statecost=0.
\end{equation}
For known dynamics $(A,B)$, it is well known that the gain
\begin{equation}\label{eq:gain}
	K(\Theta)=-\actioncost^{-1}B^\top P(\Theta)
\end{equation}
stabilizes the system. However, when the underlying dynamics are unknown, one must \emph{learn} them before a controller based on the above Riccati equation can be synthesized. Our objective is to design the control input in \eqref{eq:sde} to provide sufficient information for identification, so that the estimate $\sampleparameter$ is \emph{sufficiently close} to the true parameter $\extendedmatrix$. This estimate is then supplied to \eqref{eq:care}--\eqref{eq:gain} to obtain a gain $K(\sampleparameter)$ intended to stabilize the true system.
Before proceeding, we impose the following controllability condition on the true system.
\begin{assumption}\label{assumption 1}
	There exists a constant $\kappa>0$ such that
	\begin{equation}\label{ineq for assumption 1}
		\int_{0}^{1}
		e^{\dynamicA t}
		\controlB
		\controlB^\top
		e^{\dynamicA^\top t}
		dt
		\succeq \kappa\mathbb{I}_{\dimstate}.
	\end{equation}
\end{assumption}

The above assumption has two implications. First, by proper design of the control input, one can excite the state trajectory in every direction, which facilitates accurate identification of the system matrices. Second, it ensures that the true system is stabilizable, so sufficiently accurate estimates can be used to construct a stabilizing feedback gain.

%% file: sections/03_identification.tex
\section{System Identification for Stabilization}\label{sec:identification}
\begin{figure*}[t]
\centering
\begin{minipage}[b]{.48\textwidth}\centering
{\small $T\lVert\sampleparameter-\extendedmatrix\rVert_F^2$ vs.\ $T$ (fixed $N$)}\\[1pt]
\includegraphics[width=\linewidth]{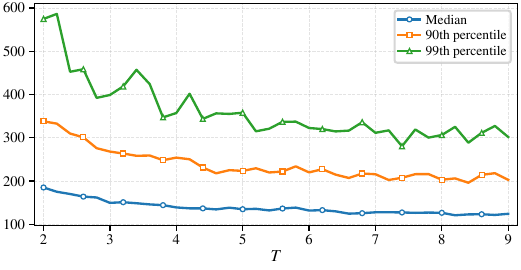}\\[-1pt]{\small (a)}
\end{minipage}\hfill
\begin{minipage}[b]{.48\textwidth}\centering
{\small $N\lVert\sampleparameter-\extendedmatrix\rVert_F^2$ vs.\ $N$ (fixed $T$)}\\[1pt]
\includegraphics[width=\linewidth]{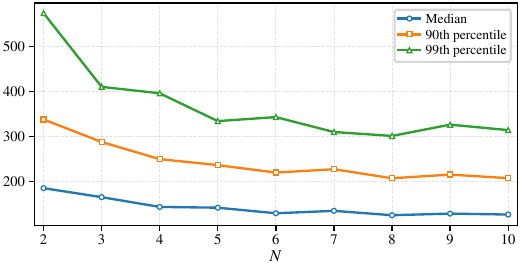}\\[-1pt]{\small (b)}
\end{minipage}
\caption{Scaled squared identification error for the system given in \eqref{eq:plant}: (a) versus $T$ at $N=2$ and (b) versus $N$ at $T=2$. Curves show the median, 90th, and 99th percentiles over $500$ runs.}\label{fig:identification}
\end{figure*}
\begin{figure*}[t]
\centering
\begin{minipage}[b]{.48\textwidth}\centering
{\small $\log\mathbb{P}(\mathcal{E})$ vs.\ $T$ (fixed $N$)}\\[1pt]
\includegraphics[width=\linewidth]{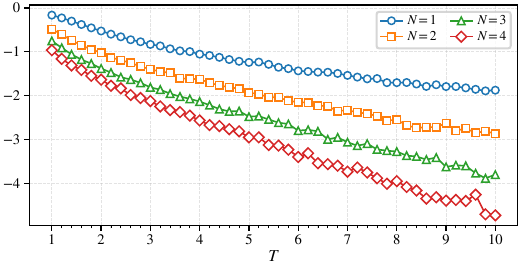}\\[-1pt]{\small (a)}
\end{minipage}\hfill
\begin{minipage}[b]{.48\textwidth}\centering
{\small $\log\mathbb{P}(\mathcal{E})$ vs.\ $N$ (fixed $T$)}\\[1pt]
\includegraphics[width=\linewidth]{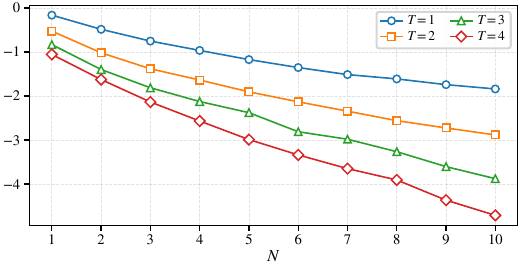}\\[-1pt]{\small (b)}
\end{minipage}
\caption{Logarithm of the stabilization failure probability for the system given in \eqref{eq:plant}, computed over $10\,000$ runs per configuration. The failure probability decreases exponentially with both (a) the trajectory length $T$ and (b) the number of trajectories $N$.}\label{fig:failure}
\end{figure*}
\begin{figure*}[t]
	\centering
	\begin{minipage}[b]{.48\textwidth}\centering
		{\small Failure probability level sets in the $(T,N)$ plane}\\[1pt]
		\includegraphics[width=\linewidth]{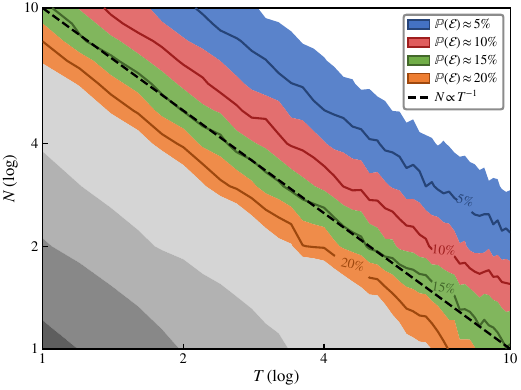}\\[-1pt]{\small (a)}
	\end{minipage}\hfill
	\begin{minipage}[b]{.48\textwidth}\centering
		{\small $\log\mathbb{P}(\mathcal{E})$ vs.\ $1/(\dimstate+\dimaction)$}\\[1pt]
		\includegraphics[width=\linewidth]{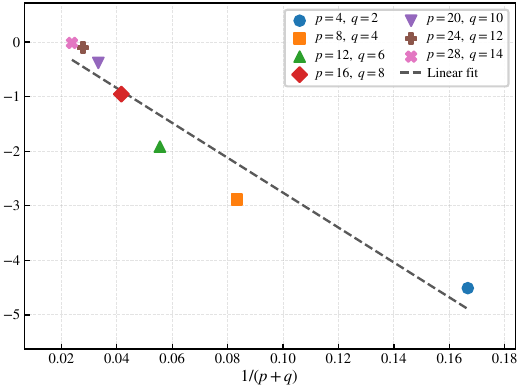}\\[-1pt]{\small (b)}
	\end{minipage}
	\caption{(a) Level sets of $\mathbb{P}\left(\mathcal{E}\right)$ for the system in \eqref{eq:plant} on log--log axes, computed over $10\,000$ runs per configuration. The dashed reference $N=10/T$ is a constant-$NT$ curve, indicating that failure probability is governed by $NT$. (b) $\log\mathbb{P}(\mathcal E)$ versus $1/(\dimstate+\dimaction)$ for $m=1,\dots,7$ block-diagonal copies of \eqref{eq:plant} at $N=10$, $T=2$, and $\sigma_{\mathbb{U}}^2=2$, computed over $2000$ runs.}\label{fig:budget}
\end{figure*}
\begin{figure*}[t]
	\centering
	\begin{minipage}[b]{.48\textwidth}\centering
		{\small $\log\mathbb{P}(\mathcal{E})$ vs.\ $1/\sigma_{\mathbb{W}}^2$}\\[1pt]
		\includegraphics[width=\linewidth]{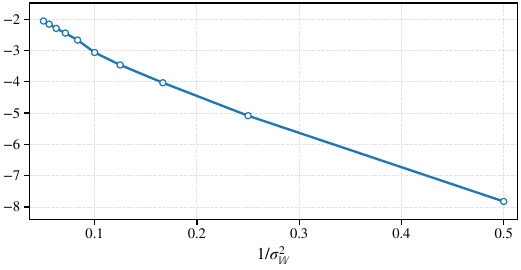}\\[-1pt]{\small (a)}
	\end{minipage}\hfill
	\begin{minipage}[b]{.48\textwidth}\centering
		{\small $\log\mathbb{P}(\mathcal{E})$ vs.\ $1+\sigma_{\mathbb{U}}^2$ (log scale)}\\[1pt]
		\includegraphics[width=\linewidth]{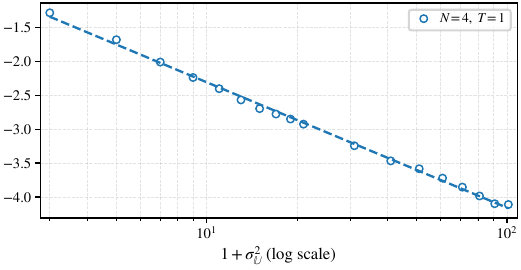}\\[-1pt]{\small (b)}
	\end{minipage}
	\caption{Logarithm of the stabilization failure probability for \eqref{eq:plant} versus (a) the inverse noise variance $1/\sigma_{\mathbb{W}}^2$ at $N=4$, $T=2$, and $\sigma_{\mathbb{U}}^2=2$, over $5000$ runs, and (b) the excitation variance $1+\sigma_{\mathbb{U}}^2$ at $N=4$ and $T=1$.}\label{fig:noise-excitation}
\end{figure*}
\begin{figure*}[t]
	\centering
	\begin{minipage}[b]{.48\textwidth}\centering
		{\small $\log\mathbb{P}(\mathcal{E})$ vs.\ $\kappa$}\\[1pt]
		\includegraphics[width=\linewidth]{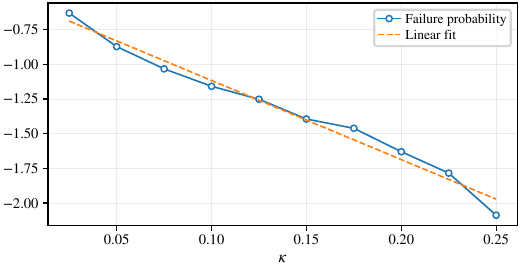}\\[-1pt]{\small (a)}
	\end{minipage}\hfill
	\begin{minipage}[b]{.48\textwidth}\centering
		{\small $\log\mathbb{P}(\mathcal{E})$ vs.\ $NT\kappa\log(1+\sigma_{\mathbb{U}}^2)/[\sigma_{\mathbb{W}}^2(\dimstate+\dimaction)]$}\\[1pt]
		\includegraphics[width=\linewidth]{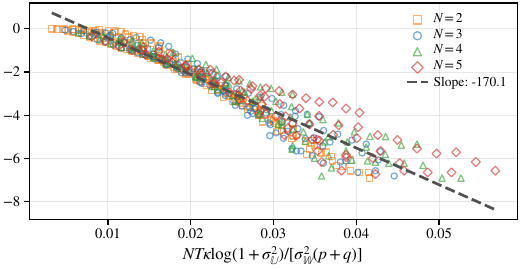}\\[-1pt]{\small (b)}
	\end{minipage}
	\caption{Logarithm of the stabilization failure probability versus (a) the controllability level $\kappa$ of the modified actuators in \eqref{eq:unstablectrl} at $N=40$ and $T=2$, over $500$ runs, and (b) the scaling variable $NT\kappa\log(1+\sigma_{\mathbb{U}}^2)/[\sigma_{\mathbb{W}}^2(\dimstate+\dimaction)]$ of \eqref{eq:rate} for two block-diagonal copies of \eqref{eq:plant} ($\dimstate+\dimaction=12$, $\kappa = 0.28$, $\sigma_{\mathbb{W}}=4$), with $N=2,\ldots,5$, $T\in[1,5]$, and $\sigma_{\mathbb{U}}^2\in[2,50]$.}\label{fig:controllability-joint}
\end{figure*}
\begin{figure*}[t]
\centering
\begin{minipage}[b]{.48\textwidth}\centering
{\small Identification error and $\mathbb{P}(\mathcal E)$ vs.\ $\kappa_s$}\\[1pt]
\includegraphics[width=\linewidth]{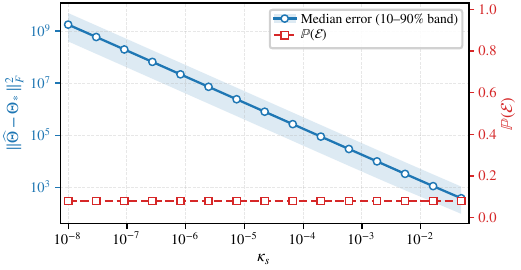}\\[-1pt]{\small (a)}
\end{minipage}\hfill
\begin{minipage}[b]{.48\textwidth}\centering
{\small $\log\mathbb{P}(\mathcal E)$ vs.\ $T$ (order misspecification)}\\[1pt]
\includegraphics[width=\linewidth]{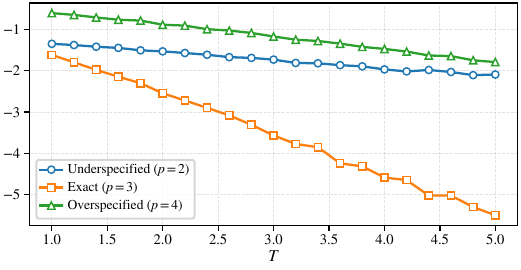}\\[-1pt]{\small (b)}
\end{minipage}
\caption{(a) Stable-mode experiment \eqref{eq:weakctrl} at $N=10$, $T=2$, over $500$ runs: median squared error with 10--90\% band (left axis) and failure probability (right axis) versus $\kappa_s$ in \eqref{eq:kappa}. (b) Order misspecification for \eqref{eq:order} at $N=2$, assumed orders $\dimstate=2,3,4$.}\label{fig:actuator}
\end{figure*}
In this section, we first describe a method that learns to stabilize \eqref{eq:sde} without prior knowledge of the dynamics or a stabilizing gain, and then present the scaling relation that organizes our experiments. Consider $N$ independent realizations of \eqref{eq:sde}, each observed over $[0,T]$. The method also applies to a single trajectory by taking $N=1$. Divide $[0,T]$ into $\numberofpieces$ subintervals
$I_j=[(j-1)\lengthofdiscrete,j\lengthofdiscrete)$
of length $\lengthofdiscrete$, so that
$T=\numberofpieces\lengthofdiscrete$.
For each realization $i=1,\ldots,N$, and subinterval indexed by $j=1,\ldots,\numberofpieces$, let
$\normalinput_j^{(i)}\sim\mathbb N(0,\mathbb{I}_{\dimaction})$
be mutually independent Gaussian vectors, also independent of the disturbing Wiener processes. The input is held constant on each subinterval:
\begin{equation}\label{eq:input}
	\actionU_t^{(i)}=\sigma_{\mathbb{U}}\normalinput_j^{(i)},
	\qquad\text{for}\qquad t\in I_j,
\end{equation}
where $\sigma_{\mathbb{U}}^2>0$ is the excitation variance. Under \eqref{eq:input}, the solution of \eqref{eq:sde} for each realization initiated at $\state_0^{(i)}$ is
\begin{equation}\label{eq:solution}
	\begin{split}
		\state_t^{(i)}={}&e^{\dynamicA t}\state_0^{(i)}
		+\sigma_{\mathbb{U}}\sum_{j=1}^{\numberofpieces}\int_{I_j\cap[0,t]}e^{\dynamicA(t-r)}\,dr\,\controlB\normalinput_j^{(i)}\\
		&+\int_0^te^{\dynamicA(t-r)}C_r\,d\WeinerPP_r^{(i)},
	\end{split}
\end{equation}
with an excitation term proportional to $\sigma_{\mathbb{U}}$ and a noise term governed by $\sigma_{\mathbb{W}}$ in \eqref{eq:upper-bound-C}. At the end of the identification interval $[0,T]$, we pool the data from all $N$ realizations to estimate $\extendedmatrix^\top=[\dynamicA\;\controlB]$ by
\begin{equation}\label{eq:ls}
	\sampleparameter=V^{-1}\sum_{i=1}^N\int_0^T
	\begin{bmatrix}\state_t^{(i)}\\ \actionU_t^{(i)}\end{bmatrix}
	d(\state_t^{(i)})^\top,
\end{equation}
where $V$ is the aggregate state--input sample covariance matrix
\begin{equation}\label{eq:Sigma}
	V=\sum_{i=1}^N\int_0^T
	\begin{bmatrix}\state_t^{(i)}\\ \actionU_t^{(i)}\end{bmatrix}
	\begin{bmatrix}\state_t^{(i)}\\ \actionU_t^{(i)}\end{bmatrix}^{\!\top}dt.
\end{equation}
The estimate in \eqref{eq:ls} can be viewed as a continuous-time least-squares estimator. It is obtained by sampling the trajectory on a mesh of size $h$, fitting the corresponding least-squares estimator, and taking the limit as $h$ tends to zero.

The estimate is then supplied to \eqref{eq:care}--\eqref{eq:gain} to obtain $K(\sampleparameter)$. Applying the feedback $\actionU_t=K(\sampleparameter)\state_t$ gives the true closed-loop matrix $\dynamicA+\controlB K(\sampleparameter)$.
We define the failure event
\begin{equation}\label{eq:event}
	\mathcal E=
	\bigl\{\text{ $\dynamicA+\controlB K(\sampleparameter)$ is not stable}\bigr\}.
\end{equation}

We study numerically how $\mathbb P(\mathcal E)$ depends on the trajectory length $T$, the number of realizations $N$, the dimension $\dimstate+\dimaction$, the noise magnitude $\sigma_{\mathbb{W}}$ in \eqref{eq:upper-bound-C}, the excitation variance $\sigma_{\mathbb{U}}^2$, and the controllability level $\kappa$ in Assumption~\ref{assumption 1}. The experiments are organized around the scaling relation
\begin{equation}\label{eq:rate}
	\log\mathbb P(\mathcal E)
	=
	-c\,\frac{NT\kappa\log(1+\sigma_{\mathbb{U}}^2)}
	{\sigma_{\mathbb{W}}^2(\dimstate+\dimaction)},
\end{equation}
where $c>0$ depends only on $\dynamicA$ and $\lengthofdiscrete$. By \eqref{eq:rate}, the failure probability decays exponentially with the aggregate observation time $NT$, so fewer, longer trajectories can achieve similar stabilization performance to that of many short ones of the same total length, despite the strong dependence along each unstable trajectory. The rate also scales inversely with the dimension and the noise variance $\sigma_{\mathbb{W}}^2$, and grows logarithmically with $1+\sigma_{\mathbb{U}}^2$.

This logarithmic dependence reflects signal saturation. Intuitively, by \eqref{eq:solution}, increasing $\sigma_{\mathbb{U}}$ amplifies the excitation term relative to the noise, which strengthens the signal. However, on each $I_j$ this term contains the response to the current input, so as $\sigma_{\mathbb{U}}$ grows, $\state_t^{(i)}$ and $\actionU_t^{(i)}$ become increasingly correlated and provide little additional information for jointly estimating $\dynamicA$ and $\controlB$. Such saturation does not arise in \citet{hafshejani2025effect}, where only $\dynamicA$ is estimated.

%% file: sections/04_numerics.tex
\section{Numerical Experiments}\label{sec:numerics}

We consider the linearized model of an unstable batch reactor used as a benchmark in networked and data-driven control with noise-free or bounded-noise data \citep{walsh2001scheduling,depersis2020formulas,rotulo2020data}. We use the continuous-time matrices reported by \citet{antunes2015stochastic} and add the disturbing Wiener process with the noise matrix $C_t=4\mathbb{I}_4$. This system is chosen for illustration only; similar results are obtained for other unstable dynamics. Values below are displayed to two decimal places:
\begin{equation}\label{eq:plant}
	\resizebox{.88\columnwidth}{!}{$\displaystyle
		\dynamicA=\begin{bmatrix}
			1.38&-0.21&6.72&-5.68\\
			-0.58&-4.29&0.00&0.68\\
			1.07&4.27&-6.65&5.89\\
			0.05&4.27&1.34&-2.10
		\end{bmatrix},\quad
		\controlB=\begin{bmatrix}
			0.00&0.00\\
			5.68&0.00\\
			1.14&-3.15\\
			1.14&0.00
		\end{bmatrix}.$}
\end{equation}
The eigenvalues of $\dynamicA$ are approximately $1.99$, $0.06$, $-5.06$, and $-8.67$. Thus, there are two unstable modes, and the state magnitude reaches the order of $10^{10}$ within approximately $10$ seconds, making fast stabilization essential.

We vary the number of trajectories $N=1,\ldots,10$ and their length $T=1,1.2,\ldots,10$. Unless stated otherwise, $\sigma_{\mathbb{U}}^2=1$, $\statecost=\mathbb{I}_{\dimstate}$, and $\actioncost=\mathbb{I}_{\dimaction}$. Figure~\ref{fig:identification} reports the squared error $\lVert\sampleparameter-\extendedmatrix\rVert_F^2$ over $500$ runs per $(N,T)$ pair, scaled by $T$ at fixed $N$ in panel (a) and by $N$ at fixed $T$ in (b). The approximately constant levels show that the squared error scales inversely with the total amount of data $NT$, up to confidence factors.

For stabilization, we estimate $\mathbb P(\mathcal E)$ for each $(N,T)$ pair as the fraction of $10\,000$ independent runs in which the resulting gain fails to stabilize the system. In Figure~\ref{fig:failure}, $\log\mathbb P(\mathcal E)$ decreases linearly in both $T$ and $N$, so the failure probability decays exponentially in each, in agreement with \eqref{eq:rate}.

In Figure~\ref{fig:budget}, panel (a) shows the level sets of $\mathbb{P}(\mathcal E)$ over the $N$--$T$ grid on a log--log scale. Each band corresponds to $(N,T)$ combinations that yield approximately the same failure probability and closely follows the slope of the dashed reference line $N=10/T$. Hence, fewer, longer trajectories perform comparably to many short ones, and extending the horizon remains informative despite the statistical dependence that instability induces along each trajectory.

To vary the dimension, we construct $m=1,\ldots,7$ block-diagonal copies of the matrices in \eqref{eq:plant}, so that $\dimstate=4m$ and $\dimaction=2m$. The estimator fits the full matrices without exploiting the block structure. Figure~\ref{fig:budget}(b) shows that $\log\mathbb{P}(\mathcal E)$ decreases approximately linearly with $1/(\dimstate+\dimaction)$ for fixed values of $(N,T)$, in agreement with the inverse dependence of the rate in \eqref{eq:rate} on dimension. Other $(N,T)$ pairs display similar behavior.

Figure~\ref{fig:noise-excitation}(a) examines noise intensity by setting $C_t=\sigma_{\mathbb{W}}\mathbb{I}_4$ and varying $\sigma_{\mathbb{W}}^2\in\{2,4,\ldots,20\}$. The decrease of $\log\mathbb{P}(\mathcal E)$ with $1/\sigma_{\mathbb{W}}^2$ shows that stabilization becomes more challenging as the noise intensity increases.

Figure~\ref{fig:noise-excitation}(b) examines the effect of the excitation magnitude $\sigma_{\mathbb{U}}^2$ at fixed noise $C_t=4\mathbb{I}_4$. Although stronger excitation reduces the failure probability, $\log\mathbb{P}(\mathcal E)$ decreases only logarithmically with $\sigma_{\mathbb{U}}^2$, in agreement with \eqref{eq:rate}. This behavior reflects the signal saturation discussed in Section~\ref{sec:identification}. Similar behavior is observed for other $(N,T)$ pairs.

To examine controllability, let $\lambda_u=0.06$ be the smaller unstable eigenvalue of $\dynamicA$, with right and left eigenvectors $v_u$ and $w_u$ such that $w_u^\top v_u=1$. We modify the input matrix as
\begin{equation}\label{eq:unstablectrl}
 B_{\delta,u}=\bigl[\mathbb{I}_4-(1-\delta)\Pi_u\bigr]\controlB,
 \qquad \Pi_u=v_uw_u^\top.
\end{equation}
Then $w_u^\top B_{\delta,u}=\delta w_u^\top\controlB$, so decreasing $\delta$ weakens only the selected mode's input coupling. We choose $\delta$ so that the smallest eigenvalue $\kappa$ of the controllability Gramian over $[0,1]$ takes the values $0.025,0.05,\ldots,0.25$. Figure~\ref{fig:controllability-joint}(a) shows that $\log\mathbb P(\mathcal E)$ decreases approximately linearly in $\kappa$, as in \eqref{eq:rate}.

Figure~\ref{fig:controllability-joint}(b) examines these effects jointly. For $m=2$ block-diagonal copies of \eqref{eq:plant}, with $\dimstate+\dimaction=12$, noise intensity $\sigma_{\mathbb{W}}=4$, and $\kappa=0.28$, it plots $\log\mathbb{P}(\mathcal E)$ for all combinations of $N$, $T$, and $\sigma_{\mathbb{U}}^2$ against $NT\kappa\log(1+\sigma_{\mathbb{U}}^2)/[\sigma_{\mathbb{W}}^2(\dimstate+\dimaction)]$, the quantity appearing in \eqref{eq:rate}. The fitted slope gives the estimate $c=170.1$ in \eqref{eq:rate} for this system.

\subsection{Identification and stabilization}\label{sec:identification-stabilization}
Next, we examine the distinction between accurate identification and successful stabilization by weakening excitation of a stable mode. Let $\lambda_s=-5.06$ be a stable eigenvalue of $\dynamicA$, with right and left eigenvectors $v_s,w_s$ satisfying $w_s^\top v_s=1$, and let $\Pi_s=v_sw_s^\top$. We set
\begin{equation}\label{eq:weakctrl}
	C_t=4(\mathbb{I}_4-\Pi_s),\qquad\text{and}\qquad B_{\delta,s}=\bigl[\mathbb{I}_4-(1-\delta)\Pi_s\bigr]\controlB,
\end{equation}
with $\delta\in(0,1]$. Then $w_s^\top C_t=0$ for all $t$ and $w_s^\top B_{\delta,s}=\delta\,w_s^\top\controlB$. Thus the noise does not excite the selected stable mode, and decreasing $\delta$ weakens its input coupling while preserving the coupling to the unstable modes. We quantify controllability by
\begin{equation}\label{eq:kappa}
	\kappa_s(\delta)=\lambda_{\min}\!\left(\int_0^1e^{\dynamicA t}B_{\delta,s} B_{\delta,s}^\top e^{\dynamicA^\top t}\,dt\right),
\end{equation}
and choose $\delta$ such that $\kappa_s$ takes $15$ logarithmically spaced values between $10^{-8}$ and $0.049$. We use $N=10$, $T=2$, and $500$ runs per configuration. The input and noise realizations are shared across the modified input matrices. 

Figure~\ref{fig:actuator}(a) reports the median squared estimation error with its 10--90\% percentile band (left axis) and the failure probability (right axis). As $\kappa_s$ decreases, the stable mode is excited by neither the noise nor the input, and its dynamics become increasingly difficult to identify. Nevertheless, stabilization failure remains constant over nearly seven orders of magnitude of $\kappa_s$. Since the weakly excited mode is stable, errors in its estimate barely affect closed-loop stability; insufficient excitation can thus degrade identification without impairing stabilization.

\subsection{Order misspecification}\label{sec:order}
Higher-order differential equations can be represented as first-order systems by augmenting the state with successive derivatives \citep{boyce2017elementary}. If the order is misspecified, estimation and feedback synthesis are performed using a state representation of incorrect dimension. We examine this mismatch using the third-order oscillator of \citet{moore1966thermally},
\[
 \dddot x+\ddot x+(\mu-R+Rx^2)\dot x+\mu x=u.
\]
Linearizing around the origin with $\mu=2$ and $R=15$, setting $\state=[x,\dot x,\ddot x]^\top$, and adding a stochastic disturbance gives
\begin{equation}\label{eq:order}
 A_o=\begin{bmatrix}0&1&0\\0&0&1\\-2&13&-1\end{bmatrix},\quad
 B_o=\begin{bmatrix}0\\0\\1\end{bmatrix},\quad
 C_o=\begin{bmatrix}1\\0\\0\end{bmatrix},
\end{equation}
with $\Re\{\lambda_{\max}(A_o)\}=3.05$.

We compare three representations of the same system. Exact specification uses $Y^{[3]}=\state$. Underspecification uses $Y^{[2]}=[x,\dot x]^\top$, omitting the highest-order state needed to represent \eqref{eq:order} exactly. Overspecification introduces $\dot z=x-z$ and uses $Y^{[4]}=[\state^\top,z]^\top$, adding a redundant dynamical coordinate without altering the system. In each case, estimation \eqref{eq:ls} and synthesis \eqref{eq:care}--\eqref{eq:gain} use the selected representation, while the resulting feedback is applied to the true system and evaluated through \eqref{eq:event}. Figure~\ref{fig:actuator}(b) shows that exact specification gives the fastest decay of the failure probability, while both mismatches substantially degrade the stabilization performance.

%% file: sections/05_deficiencies.tex
\section{Concluding Remarks}\label{sec:conclusion}
We numerically study identification for stabilization of unknown continuous-time stochastic linear systems. Stabilization performance improves with the aggregate observation time, excitation, and controllability, and degrades with noise intensity and dimension, in agreement with \eqref{eq:rate}. In particular, despite instability, a single long trajectory performs comparably to many short trajectories of the same total length. We also observe signal saturation: stronger excitation helps, but with diminishing returns, since the logarithm of the failure probability depends only logarithmically on the excitation variance. Moreover, accurate identification of the full dynamics is not necessary for stabilization, which suggests that identifying the unstable modes may suffice and offers favorable data requirements for systems with few unstable and many stable modes. Future work includes a theoretical analysis of \eqref{eq:rate}, a study of the sufficiency of unstable-mode identification for stabilization, and extensions to time-varying dynamical systems.